\documentclass{article}
\usepackage{ijcai16}
\usepackage{algorithm}
\usepackage{algpseudocode}
\usepackage{pifont}
\usepackage{booktabs}
\usepackage{multirow}
\usepackage{graphicx}
\usepackage{epstopdf}
\usepackage{amssymb,amsthm,bbm,amsmath}

\usepackage{times}

\newtheorem{thm}{Theorem}
\newtheorem{cor}{Corollary}

\newtheorem{lemma}{Lemma}
\theoremstyle{definition}

\theoremstyle{remark}

\newcommand{\bw}{\mathbf{w}}

\newcommand{\bx}{\mathbf{x}}
\newcommand{\by}{\mathbf{y}}

\newcommand{\tr}{\mathbf{T}}
\newcommand{\bX}{\mathbf{X}}
\newcommand{\bXt}{\mathbf{X}_t}

\newcommand{\bB}{\mathbf{B}}
\newcommand{\bbt}{\mathbf{b}_t}
\newcommand{\bb}{\mathbf{b}}
\newcommand{\bP}{\mathbf{P}}

\newcommand{\cB}{\mathcal{B}}
\newcommand{\cD}{\mathcal{D}}

\newcommand{\reals}{\mathbb{R}}

\newcommand{\nr}{\mathbb{R}}

\newcommand{\eqdef}{\triangleq}

\newcommand{\EP}{\textbf{EREP}}

\newcommand{\Gcite} [1] {\citeauthor{#1} \shortcite{#1}}

\title{Online Learning of Portfolio Ensembles with Sector Exposure Regularization}
\author{Guy Uziel  \\ 
Technion -- Israel Institute of Technology  \\
Ran El-Yaniv  \\ 
Technion -- Israel Institute of Technology  \\
}

\begin{document}

\maketitle

\begin{abstract}
We consider online learning of ensembles of portfolio selection algorithms
and aim to regularize risk by encouraging diversification with respect to  a predefined 
risk-driven grouping of stocks.
Our procedure uses online convex optimization to control capital 
allocation to underlying investment algorithms while encouraging non-sparsity over
the given grouping.
We prove a logarithmic regret for this procedure with respect to the 
best-in-hindsight ensemble. We applied the procedure with known 
mean-reversion portfolio selection algorithms using the standard GICS industry sector grouping. Empirical Experimental results
showed an impressive percentage increase of risk-adjusted return (Sharpe ratio).
\end{abstract}

\section{Introduction}
\label{sec:intro}
Online \emph{portfolio selection} \cite{Cover1991} has become one of the focal points in online 
learning research.
So far, papers along this line have mainly considered \emph{cumulative wealth} or \emph{return}
as the primary quantity to be optimized. However, 
the quality of any investment should rather be quantified using both
return and \emph{risk}, which often is measured by the variance of the return.\footnote{Many other notions of 
	risk have been considered as well; see, e.g. \Gcite{Harlow1991}.}
 One of the major open challenges in online portfolio selection is 
the incorporation of effective mechanisms to dynamically control \emph{risk}  \cite{LiH2014}.
Within a \emph{regret minimization} framework, like the one we consider here, 
this challenge is highlighted by the impossibility of achieving 
sub-linear regret in the adversarial setting
with respect to risk adjusted measures such as the Sharpe Ratio \cite{EvenKW2006}.
This theoretical limitation is perhaps the main reason 
for the scarcity of papers studying risk control in the context 
of online portfolio selection.\footnote{This lack of risk-aware results in online (adversarial) portfolio learning 
	stands in stark contrast to the situation in classic finance where the research into risk is vast  within the context of 
	portfolio allocation under distributional assumptions.}  

Following \Gcite{JohnsonB2015}
and
motivated by the classic idea of diversifying risk factors and in particular, 
industry sectors (e.g., \Gcite{Grinold1989}, \Gcite{CavagliaBA2000}),
in this paper we present 
a portfolio selection ensemble learning procedure that 
applies any set of  portfolio selection algorithms, and 
controls risk using sector \emph{regularization}.
The procedure invests in the stock market by dynamically allocating capital among the  
base algorithms, which provide their investment recommendations.
These dynamic allocations are online learned using Newton steps, and
risk is controlled using an  $\ell_{\infty} / \ell_1$ 
 regularization term that penalizes over-exposed portfolios,
where exposure is defined and quantified using a
prior grouping of the stocks into ``industry sectors.''
While this regularization 
elicits diversification among groups,  
it also encourages a focus on the best convex combination 
of base algorithms within groups.
The set of base trading algorithms can be arbitrary, but a typical application might be to activate
within groups
several independent copies of the same base algorithm
instantiated
with different values of its hyper-parameters, thus solving or alleviating the challenge of
hyper-parameter tuning.

We prove a logarithmic regret bound for our procedure 
with respect to the best-in-hindsight ensemble of the base algorithms.
We also show preliminary, promising 
numerical examples over a commonly used and challenging benchmark dataset.
These results demonstrate impressive improvements in risk-adjusted return (Sharpe ratio)
relative to direct applications of the base algorithms, and compared to a  previous 
attempt to utilize group diversification.

 \subsection{Online Portfolio Selection }
 \label{sec:OPS}
 In Cover's classic portfolio selection setting \cite{Cover1991} (see also \Gcite{Borodin2005}, Chapt. 14),
 one assumes a market of $n$ stocks. 
 We consider an online game played through $T$ days.
 On each day $t$ the market is represented by
 a \emph{market vector} $\bXt$ of relative prices, 
 $\bXt \eqdef (x_{1}^{t},x_{2}^{t},...,x_{n}^{t})$,
 where for each $i=1, \ldots ,n$, 
 $x_{i}^{t}\geq0$ is the \emph{relative price} of stock $i$, defined to be the ratio of its closing price on day $t$ relative to its closing price on day $t-1$. 
 We denote by $\bX \eqdef \bX_{1}, \ldots, \bX_{T}$
 the sequence of $T$ market vectors for the entire game.
 A \emph{wealth allocation} vector or \emph{portfolio} for day $t$ is 
 $\bbt \eqdef (b_{1}^{t}, b_{2}^{t} , \ldots, b_{n}^{t})$, where
 $b_{i}^{t}\geq0$ is the wealth allocation for stock $i$.
 We require that the portfolio satisfy $\sum_{i=1}^{m}b_{i}^{t}=1$. 
 Thus, $\bbt$ specifies the online player's 
 wealth allocation for each of the $n$ stocks on day $t$, and $b_{i}^t$ is the fraction
 of total current wealth invested in stock $i$ on that day.
 We denote by $\bB \eqdef  \bb_1, \ldots , \bb_T$ the sequence of
 $T$ portfolios for the entire game. 
 At the  start of each trading day $t$, the player
 chooses a portfolio $\bb_t$. Thus, by the end of day
 $t$, the player's wealth is multiplied by 
 $\left\langle \bb_t ,\bXt \right\rangle = \sum_{i=1}^{n}b_i^t x_i^t$,
 and assuming initial wealth of \$1, the player's cumulative wealth by the end of the game
 is therefore
 
 \begin{equation}
 R_T (\bB, \bX)  \eqdef  \prod_{t=1}^{T}
 \left\langle \bbt ,\bXt \right\rangle .
 \end{equation}
 
 In the setting above, it is common to consider the logarithmic cumulative wealth, $\log R_T (\bB, \bX) $, which can be expressed as 
 a summation of the logarithmic daily wealth increases, $\log(\left\langle \bbt ,\bXt \right\rangle )$.
 
 The basic portfolio selection problem as defined above 
 abstracts away various practical considerations, such as 
 commissions, slippage, and more generally, market impact, which are crucial for 
 realistic implementations. While we do ignore these elements here as well, 
 we focus in the present work on  \emph{risk},
 and adapt the traditional definition of risk as the variance of return.
 Let $R_A$ be the (annualized) return of an investment algorithm $A$ and let $\sigma_A$ be the 
 (annualized) standard deviation of $A$'s return. Let $R_f$ be the risk-free (annualized) interest rate. 
 Then the (annualized) Sharpe ratio \cite{Sharpe1966} of $A$ is    
 \begin{equation*}
 \label{eq:Sharp}
 S =\frac{R_A - R_f}{\sigma_A} .
 \end{equation*}
 For comparative purposes it is common to ignore the risk-free return $R_f$ as we do here.
 The Sharpe ratio is thus a measure of risk-adjusted return, which captures the expected differential return per unit of risk.

 
 In the online (worst-case) approach to portfolio learning the goal is to online
 generate a sequence $\{ \bb_t \}$ of portfolios that compete with the best-in-hindsight fixed portfolio,
 denoted $\bb_*$. Denoting the round $t$ loss of portfolio $\bb$ by $f_t(\bb)$ (in our case, 
 $f_t(\bb) = -\log(\left\langle \bbt ,\bXt \right\rangle )$), we define the regret of sequence 
 $\{ \bb_t \}$ as
$$ 
\textbf{Regret} \eqdef \sum^T_{t=1} \left(f_t(\bb_t)-f_t(\bb_*)\right) .
$$
In this paper we are mainly concerned with portfolio \emph{ensembles}, where the allocation $\bb_t$
is over trading algorithms and $\bb_*$ represents the optimal-in-hindsight fixed allocation.

\subsection{Risk Reduction by Sector Regularization}
\label{sec:onDiversification}

Diversification is the process of allocating wealth among investment choices 
such that exposure to certain ``risk factors''
is controlled or reduced. 
Markowitz's modern portfolio theory (MPT) put diversification on front stage
by showing a systematic diversification procedure for static portfolios using correlation analysis \cite{Markowitz1952}.
Understanding financial risk factors, and their interrelationship with stock returns, has been a longstanding challenge:
one of the profound insights has been that risk factors
can be manifested in many ways, but   not all factors can be diversified
\cite{Sharpe1964}.
Among the well-known diversifiable factors are those which are
country specific and industry related \cite{Grinold1989}.
Several decades ago diversification across countries provided greater risk reduction than 
industry-wise diversification \cite{Solnik1995}, but with increasing economic globalization, industry sector diversification  has been found to be of increasing importance to active portfolio management \cite{CavagliaBA2000}.

The  procedure proposed in this paper can, in principle, handle many types of diversification
expressed in terms of predefined groupings of the stocks. While computation of effective groupings with sufficient predictive power is an interesting topic in and of itself, here
we treat the grouping itself 
as available \emph{prior knowledge}.
Thus, for concrete validation of the  proposed procedure, we focus on industry sector diversification, and our numerical examples make use of 
the \emph{global industry classification standard} (GICS) -- an industry taxonomy developed in 1999 by MSCI and Standard \& Poor (S\&P) for use by the global financial community.\footnote{The GICS structure consists of 10 sectors, 24 industry groups, 67 industries and 156 sub-industries.}


\subsection{Related Work and Contributions}
\label{sec:Related}

We focus on online learning  of sequential portfolios, and
the main contextual anchor of the present work is
the line of research pioneered by Cover \cite{Cover1991,CoverO1996,OrdentlichC1996},
where the vanilla online portfolio selection problem was introduced and initially studied.
Within this line the goal is to devise  portfolio selection strategies whose cumulative wealth 
achieves, under adversarial inputs,
sublinear regret (aka ``universality'') with respect to an optimal in-hindsight 
``constant rebalanced portfolio'' (CRP) strategy. 
The regret lower bound of $\Omega(\log T)$ \cite{OrdentlichC1996} 
for a $T$-round portfolio game was originally matched  by Cover's celebrated ``Universal Portfolios'' algorithm \cite{Cover1991}, and then rematched by various similar, or other  ``follow the leader'' strategies (see, e.g.,\Gcite{helmboldSSW1998}, \Gcite{BlumK1999}, \Gcite{AgarwalHKS2006}). 

One of the approaches for regret minimization in the online portfolio setting is  online convex optimization, where in each round $t$ we consider a loss function, 
$f_t(\bbt)=-\log(\left\langle \bbt ,\bXt \right\rangle )$, and exploit its convexity to ensure 
$O(\sqrt{T})$ regret from the best-in-hindsight CRP portfolio.
\Gcite{AgarwalHKS2006} showed that by using such a loss function that is
exp-concave, it is possible to guarantee an improved regret bound of $O(\log{T})$. 
This result was further extended by \Gcite{DasB2011}, who proposed MA$_{ons}$  to ``ensemble learning,'' whereby the portfolio learned is over algorithms rather than the stocks
themselves. 
This resulted in $O(\log{T})$ regret from the best-in-hindsight convex combination of algorithms.


To the best of our knowledge, there are only a few studies of risk control in the context of online portfolio learning with bounded regret.
As mentioned earlier, \Gcite{EvenKW2006} considered a more general expert setting and showed that one cannot achieve sub-linear regret with respect to the risk-adjusted return (Sharpe ratio).
As a remedy, that paper considered optimizing global return in conjunction with a locally computed Sharpe ratio
(defined over a recent historical window). 
Within a stochastic bandit setting,  \Gcite{ShenWJZ2015} considered tracking the best expert with respect to the Sharpe ratio. 
The closest work to ours is that of
  \Gcite{JohnsonB2015}, who introduced the following group norm to encourage diversification.
 %
Consider a grouping $\mathcal{G} = (g_1, . . . , g_m)$ of the integers $[n] \eqdef \{1, \ldots, n\}$,
where  $g_i \subseteq [n]$,  $|g_i| = n_i$, and the groups
$g_i$  may overlap. For a vector $\bX \in \reals^n$, define its $\ell_{\infty} / \ell_1$ \emph{group norm}
\begin{equation}
\label{eq:Groupnorm}
L^\mathcal{G}_{(\infty,1)}(\bX)= || (||\bX_1||_1, \ldots, ||\bX_m||_1) ||_\infty ,
\end{equation}
where $\bX_i \in \reals^n$ equals $\bX$, with all coordinates in $[n] \setminus g_i$ zeroed.
This hierarchical group norm, which can be viewed as a kind of inverse to the group 
norm used in group Lasso \cite{Yuan2006}, encourages \emph{non-sparsity} in its outer norm.
\Gcite{JohnsonB2015} presented an algorithm, called ORSAD, which uses this norm  
to encourage diversification in portfolio learning, thus leading 
to a reduction in several risk parameters, including a variant of the Sharpe ratio.
The ORSAD algorithm consists of  minimization steps
of the form 
\begin{equation*}
\label{eq:Banerjee}
\arg\min_{L^\mathcal{G}_{(\infty,1)}(b_t)\leq K} -\eta\log(\left\langle \bbt ,\bXt \right\rangle )+\frac{1}{2}||\bbt-\mathbf{b}_{t-1}||_2^2 ,
\end{equation*}
where $K$ is a hyper-parameter.
They also proved that their algorithm can guarantee $O(\sqrt{T})$ regret w.r.t. the best fixed portfolio in hindsight.

Although our procedure relies on the same $\ell_{\infty} / \ell_1$  group norm to encourage diversification, our algorithm differs from that of Johnson and Banerjee in two ways: first, rather than generating portfolios on the stock themselves, we generate a weighted ensemble over investment algorithms. Second, 
our learning algorithm exploits the exp-concavity of our loss function, allowing the use of online Newton steps as in \cite{AgarwalHKS2006} to guarantee $O(\log{T})$ regret w.r.t. the best fixed ensemble in hindsight. A direct application of our procedure over the stocks themselves yields exponential 
improvement in regret relative to the result of \Gcite{JohnsonB2015}. Also worth mentioning is that 
 our implementation 
utilizes a fixed grouping of the stocks given by the standard GICS industry taxonomy, whereas  \Gcite{JohnsonB2015}
employ a correlation-based heuristic to group the stocks on the fly.
Our numerical examples in Section 3 include a direct comparison with the method of
\Gcite{JohnsonB2015}, showing an overwhelming advantage to our method
(see, e.g., Figure~\ref{fig:EPvsORSAD}).

\begin{algorithm}[htb]
	\caption{Ensemble Procedure ($\EP$)}
	\label{alg:HPS}
	\begin{algorithmic}[1]
		\State Input: $d$ trading algorithms,$k$ groups, $T, \epsilon,\eta,\lambda>0$.
		\State Initialize:$\bP_1$, $\bw_1=(\frac{1}{kd}, \ldots,\frac{1}{kd}), A_{0} = \epsilon I_{kd}$
		\For{$t=1$  to  $T$}
		\State {\bf Play} $\bw_t$ and suffer loss $g_t(\bw_t)+\lambda L^\mathcal{G}_{(\infty,1)}(\mathbf{w})$
		\State {\bf Receive} portfolios $\bP_{t+1}$ of base algorithms 
		\State {\bf Update:} $A_{t}=A_{t-1}+\nabla g_{t}(\bw_t)^{\tr}
		\nabla g_{t}(\bw_t)$ and
		\begin{multline*}
		\mathbf{w}_{t+1}=\mathbf{\arg\min_{w\in \cB}\{}\left\langle \nabla g_{t}(\mathbf{w}_{t}),\mathbf{w}-\mathbf{w}_{t}\right\rangle \\ +\lambda L^\mathcal{G}_{(\infty,1)}(\mathbf{w}) +\eta \cD_{A_{t}}(\mathbf{w}||\mathbf{w}_{t}))\} 
		\end{multline*}
		\EndFor
	\end{algorithmic}
\end{algorithm}

\section{Exposure Regularized Ensemble Procedure} 
We consider a given grouping $\mathcal{G} = (g_1, . . . , g_k)$ of the integers $[n] \eqdef \{1, \ldots, n\}$,
where  $g_i \subseteq [n]$,  $|g_i| = n_i$.  
Each group  $g_i$ is called a \emph{sector}.
We assume that the sectors represent a meaningful
 structure of the $n$ stocks and are not concerned here with how it is computed.

Our \emph{exposure regularized ensemble procedure} (henceforth, $\EP$) is constructed over a set of 
$d$ base-algorithms $\mathbf{A}=A_1, \ldots ,A_d $.
For each $1 \leq i \leq d$, we create for algorithm $A_i$, $k$ sub-algorithms $A_{i,j}$, 
$j=1, \ldots ,k$,  such that $A_{i,j}$ operates only over  sector $j$. Therefore, $\EP$
effectively operates over $kd$ sub-algorithms.
In each round $t$, $\EP$ invests its current wealth in the sub-algorithms according to an allocation vector  
$\bw_t$, which resides in the probability simplex. 
The actual allocation of wealth to individual stocks is calculated using $\bw_t$,
by aggregating  the proposed portfolios by each of the sub-algorithms in a straightforward manner.

$\EP$ requires the following definitions and notation.
Let $A \in \reals^{n \times n} $ be any positive-definite matrix.
For $\bx, \bw \in \reals^n$, the \emph{Bregman divergence} generated by $F_A(\bw) \eqdef \frac{1}{2}\bw^{T}A\bw$ is
$$
\cD_A(\bw||\bx) \eqdef
\frac{1}{2}||\bw-\mathbf{x}||_A^2=\frac{1}{2}(\bw-\mathbf{x})^{T}A(\bw-\mathbf{x}).
$$
We denote by $I_n$  the unit matrix of order $n$.
For a function $f : \reals^{n}\rightarrow \reals$, we denote by $\nabla f(\bw)$ its gradient (if it is differentiable) and by $f'(\bw)$ its subgradient.

We now explain the pseudo-code of $\EP$ listed in Algorithm~\ref{alg:HPS}. 
In each round $t$,
$\EP$ first
plays (rebalances its portfolio) according to already computed allocation vector $\bw_t$ (line 4).
In response, the adversary selects a market vector (still line 4), which determines the 
following loss $g_t(\bw_t)$:
\begin{equation}
\label{eq:gt}
g_t(\bw) \eqdef -\log (\left\langle \bX_t,\mathbf{P}_t\bw\right\rangle),
\end{equation} 
where $\bX_t$ is the market vector selected by the adversary for round $t$.
$\EP$ then receives $\bP_t \in \reals^{n \times kd}$, the revised portfolios of its sub-algorithms (line 5).
In line 6, $\EP$ updates its allocation vector. 
In order to exploit the exp-concavity of the loss function, $\EP$ utilizes the curvature of the loss function, 
as embedded in the matrix $A_t$, and then uses the Bregman divergence corresponding to $A_t$
so as to optimize its allocation vector based on second order information (Newton step).
The regularization term, $L^\mathcal{G}_{(\infty,1)}(\bw)$, encourages
the simultaneous tracking of  the most profitable sub-algorithm combination in each sector as
well as diversification over sectors, as discussed in Sections~\ref{sec:onDiversification} and~\ref{sec:Related}.

\subsection{Regret Analysis}
In this section we analyze $\EP$ and prove a logarithmic regret worst case bound. 
Let $\alpha$ be a positive real. A convex function $f$ : $\reals^{n}\rightarrow \reals$ is 
$\alpha$\emph{-exp-concave} over the convex domain $\cB \subset \reals^{n}$
if the function $\exp({\text{-\ensuremath{\alpha}}f(\mathbf{x})})$ is concave.
It is well known that the class of exp-concave functions strictly contains the class of 
strongly-convex functions. 
For example, the loss function typically used in online portfolio selection, 
$f_t(\mathbf{b})=-\log(\left\langle \bb ,\bXt \right\rangle )$,
is exp-concave but not strongly-convex.


The following two (known) basic lemmas concerning 
exp-concavity will be used in the proofs 
of Lemma~\ref{lem:10-1}  and Theorems~\ref{thm:Main} that follow.

\begin{lemma} \cite{AK2007} 
	\label{lem:expConcave}
	Let $f$ be an $\alpha$-exp-concave over $\cB\subset\nr^{n}$
	with diameter $D$, such that $\forall \bx\in \cB$ , $||\nabla f(\bx)||_{2}\leq G$.
	Then, for $\eta\leq\frac{1}{2}\min\{\alpha,\frac{1}{4GD}\}$, and for
	every $\bx,\by\in \cB$,
	\begin{multline*}
	f(\by)\geq f(\bx)+\left\langle \nabla f(\bx),\by-\bx\right\rangle \\ 
	+\frac{\eta}{2}(\by-\bx)^{T}\left(\nabla f(\bx)\nabla f(\bx)^{T}\right)(\by-\bx) . 
	\end{multline*}
\end{lemma}

\begin{lemma} \cite{AK2007}
	\label{Exp-concave_hazan}
	Let $f_{t}:\nr^{n}\rightarrow\nr$ be $\alpha$-exp-concave, and  let $A_{t}$ be as in Algorithm~\ref{alg:HPS}. 
	Then, for $\eta=\frac{1}{2}\min\{\alpha,\frac{1}{4GD}\}$
	 and $\epsilon_0=\frac{1}{\eta^2D^2}$,
	\[\sum_{t=1}^{T}||\nabla f_{t}(\bw_{t})||_{A_{t}^{-1}}^{2}\leq n\log T.\]
\end{lemma}

We consider a standard online convex optimization game \cite{Zinkevich2003} where in 
each round $t$ the online player selects a point $\bw_t$ in a convex set $\cB$; then a convex payoff function $f_t$ is revealed, and the player suffers loss $f_t(\bw_t)$. 
In an adversarial setting, where $f_t$ is selected in the worst possible way,
it is impossible to guarantee absolute online performance. Instead, the objective of the online player is
to achieve sublinear regret relative to the best choice in hindsight, 
$\bw_* \eqdef \arg\min_{w\in \cB} \sum_t f_t(\bw)$, where regret is 
$$
\textbf{Regret} \eqdef \sum^T_{t=1} \left(f_t(\bw_t)-f_t(\bw_*)\right)  .
$$

The mirror descent algorithm  \cite{NemirovskyY1983,BeckT2003}
for online convex optimization
was extended by \Gcite{DuchiSST2010} as follows.
Instead of solving in each round 
$$
\mathbf{w}_{t+1}=\mathbf{\arg\min_{w\in K}\{}\eta\left\langle \nabla f_{t}(\mathbf{w}_{t}),\mathbf{w}-\mathbf{w}_{t}\right\rangle+\cD(\mathbf{w}||\mathbf{w}_{t}\mbox{)})\},
$$
where $\cD(x||y)$ is the Bregman divergence generated by some strongly convex function $\psi$,
they proposed to solve 
\begin{align*} 
\mathbf{w}_{t+1}&=\mathbf{\arg\min_{w\in K}\{}\eta\left\langle \nabla f_{t}(\mathbf{w}_{t}),\mathbf{w}-\mathbf{w}_{t}\right\rangle +\eta r(\mathbf{w}) \\ 
& +\cD(\mathbf{w}||\mathbf{w}_{t}\mbox{)})\} ,
\end{align*}
where $r(\cdot)$ is some convex function which is not necessarily smooth.
They proved that their revised method guarantees $O(\sqrt{T})$ regret relative to 
the best choice in hindsight whenever $f$ is convex.
Moreover, a sharper $O(\log(T))$ regret bound was shown for \emph{strongly convex} $f$. 
This extension, called composite objective mirror descent (COMID),  opened the door to
applications in many fields and, in particular, to the possibility 
of using an $\ell_{\infty} / \ell_1$ group norm as we do here.
Our analysis of $\EP$ thus boils down to extending the COMID framework for exp-concave functions.



We state without proof the following results (Lemma~\ref{lem:10-1} and Theorem~\ref{thm:Main}). 
Full proofs of these statements will be presented in the long version of 
this paper.


\begin{lemma}
	\label{lem:10-1}
	Let $f_t$ be $\alpha$-exp-concave over $\cB\subset\nr^{n}$
	with diameter $D$, such that $\forall \bw\in \cB$ , $||\nabla f(\bw)||_{2}\leq G$.
	If $\bw_{t}$ is the prediction of Algorithm~\ref{alg:HPS} in round $t$,
	then, for $\eta=\frac{1}{2}\min\{\alpha,\frac{1}{4GD}\}$ and for any $\bw_{*}\in \cB$, 
	$$
	\frac{1}{\eta}\left[f_{t}(\bw_{t})-f_{t}(\bw_{*})+r(\bw_{t+1})-r(\bw_{*})\right] \leq
	$$
	$$ 
	\cD_{A_{t-1}}(\bw_{*}||\bw_{t})-\cD_{A_{t}}(\bw_{*}||\bw_{t+1})+\frac{1}{2\eta^{2}}||\nabla f_{t}(\bw_{t})||_{A_{t}^{-1}}^{2} .
	$$
\end{lemma}

\begin{thm}
	\label{thm:Main} 
	Let $f_t$ be $\alpha$-exp-concave over $K\subset\nr^{n}$,
	$\eta=\frac{1}{2}\min\{\alpha,\frac{1}{4GD}\}$ and let $\epsilon_0=\frac{1}{\eta^2D^2}$.
	If $(\bw_{1},\bw_{2}, \ldots ,\bw_{T})$ are the predictions of Algorithm~\ref{alg:HPS}, then
    for any fixed point $\bw_{*}\in \cB$,
	$$
	\sum_{t=1}^{T}\left(f_{t}(\bw_{t})+r(\bw_{t}\mbox{)}-f_{t}(\bw_{*})-r(\bw_{*}\mbox{)}\right)=O(\log T).
	$$
\end{thm}
%
%
%

\begin{cor}
For Algorithm~\ref{alg:HPS}, for appropriate\footnote{In our applications we used the same parameters that were used in \Gcite{AgarwalHKS2006}. In general, these parameters can be calibrated according to market variability;
	see  \Gcite{AgarwalHKS2006}.} 
$\epsilon,\eta>0$,
and every $\lambda\geq 0$, for any fixed point $\bw_* \in \cB$, it holds that
	\begin{gather*}
	\sum_{t=1}^{T}g_{t}(\bw_{t})+\lambda L^\mathcal{G}_{(\infty,1)}(\mathbf{w})	
	-g_{t}(\bw_{*}) 
	-\lambda L^\mathcal{G}_{(\infty,1)}(\mathbf{w}_*)\\=O(\log T).
	\end{gather*}

\end{cor}

%

\section{Numerical Examples}
\label{sec:Empirical}
In this section we present a preliminary empirical study examining and analyzing
the performance of $\EP$ on the well-known SP500 benchmark 
dataset \cite{BorodinEG2000}. 
This challenging dataset consists
of 25 stocks over a period of 5 years, from 1998 to 2003, which includes the  dot-com crash.  
Qualitatively similar results to those presented here will be presented for other datasets  in the extended version of this paper.

The stocks in the SP500 dataset were categorized into the following 4 sectors according to the global industry classification standard (see, e.g., Yahoo Finance): Technology, Finance, Healthcare, and Services. 
Fixing this sector grouping throughout our study we examine, in our first experiment, how well $\EP$ controls and operates sets of base algorithms.
We selected the following set of base algorithms, all of which are implemented in the Li et al.
OLPS simulation library  \cite{OLPS}:
\begin{itemize}
	\item
	Exponentiated Gradient (EG)~\cite{helmboldSSW1998}: this classic  algorithm is among  the early universal algorithms.
	EG is typically not a strong contender in empirical studies and we include it as a control point,
	to verify that our strategy avoids using its portfolios.
	 \item
	 Anticor~\cite{BorodinETG2004}: one of the first algorithms 
	 designed to aggressively exploit mean-reversion via 
	  (anti) correlation analysis. 
	 \item
	 Online Moving Average Reversion (OLMAR)~\cite{LiH2012}: designed to exploit mean-reversion based on moving average predictions.  OLMAR is known to be a strong performer in many benchmark datasets.
\end{itemize}
 We examined two different settings for the base algorithms. 
 In the first setting (called ``Mixed''), the set of base algorithms consists of  the three algorithms: EG , Anticor and OLMAR.\footnote{All hyper-parameters of the base algorithms were 
 	set to the recommended default parameters in the OLPS simulator. EG (with $\eta=0.05$) , Anticor (with $w=20$) and OLMAR (with $w=20,\epsilon=10$).}
In the second setting (called ``Olmar only''), we took three instances of OLMAR 
applied with the following values of its window size (a critical hyper-parameter): 10, 15, 20.

The critical hyper-parameter of $\EP$ is $\lambda$, which controls the  
$\ell_{\infty} / \ell_1$ regularization intensity.
Preliminary empirical measurements of the dynamic range of the average maximal sector weight as a function of  $\lambda$ 
showed that $\lambda = 0.1$ roughly corresponds to the median of this range.
Therefore, to obtain a rough impression we initially applied $\EP$ with this setting.
In a more elaborate experiment we optimized $\lambda$ using a standard walk-forward procedure
\cite{Pardo1992}; see details below.

\begin{table*}[htb!]
	\caption{SP500 Dataset: Sharpe ratio performance  of $\EP$ and benchmark algorithms} \label{Table:HPS}
	\begin{center}
		\small
		\setlength\tabcolsep{2.5pt}
		\begin{tabular} { l |  l l l | l | l | l l } 
			{\bf Setting}  & \multicolumn{3}{c}{{\bf Base Algorithms}}& MA$_{ons}$ &ORSAD&\multicolumn{2}{c}{{\bf $\EP$}}   \\
			
			&  & & & & & $\lambda=0.1$ & $\lambda_{WF}$  \\
			\midrule
			\multirow{ 3}{*}{Mixed}
			&               EG & Anticor$_{(w=20)}$ & OLMAR$_{(w=20)}$  \\
			& $0.51$ & $0.90$ & $0.94$ &$0.91$&  $0.52$ & \bf{1.12} & \bf{1.14}\\			
			\midrule
			\multirow{ 3}{*}{Olmar only} 					
			&               $w=10$ & $w=15$ & $w=20$  \\
			& $0.90$ & $0.77$ & $0.94$ & $0.97$& $0.52$ & \bf{1.33}  & \bf{1.39}\\			
			\midrule
		\end{tabular}
	\end{center}
\end{table*}

\begin{figure}[ht]
\vskip -0.2in
\begin{center}
\centerline{\includegraphics[height=6.5cm,width=\columnwidth]{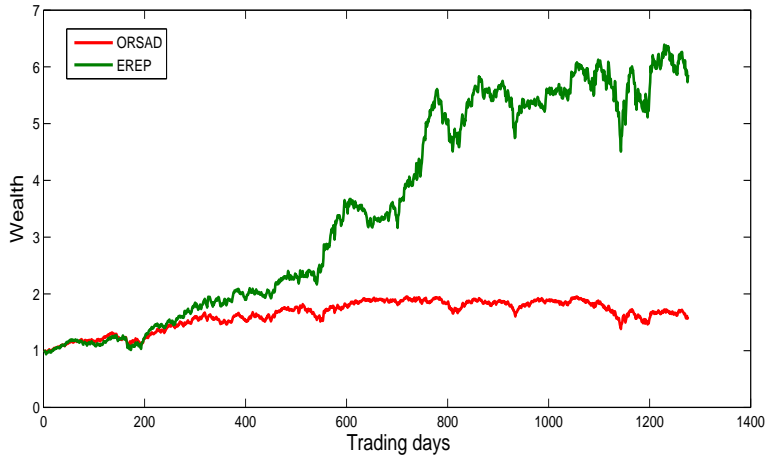}}
\caption{EREP compared to ORSAD}
\label{fig:EPvsORSAD}
\end{center}
\vskip -0.2in
\end{figure} 


Table~\ref{Table:HPS} summarizes the Sharpe ratio results obtained for these runs of 
$\EP$ compared to various benchmark algorithms.
Consider first the Mixed setting (first row of the table). Here we see
that $\EP$ improves the Sharpe ratio of the best algorithm by almost 20\%. In the Olmar only setting, the Sharpe ratio improvement relative to the best Olmar instance  is by over 40\%. 
Similar improvements can be seen with respect to the MA$_{ons}$ ensemble procedure of 
\Gcite{DasB2011}, applied here on the same sets of base algorithms.  
Turning now to the ORSAD algorithm of \Gcite{JohnsonB2015}, which uses the same 
$\ell_{\infty} / \ell_1$ regularization (but with a different learning algorithm applied over the stocks),
 we see an improvement of over  100\% 
in the Sharpe ratio. Furthermore, in 
Figure~\ref{fig:EPvsORSAD} we see the cumulative return curves of both $\EP$ and ORSAD for the
entire 5-year period, showing that the return itself has also improved by hundreds of percents.

\begin{figure}[ht!]
\vskip 0.2in
\begin{center}
\centerline{\includegraphics[height=6.5cm ,width=\columnwidth]{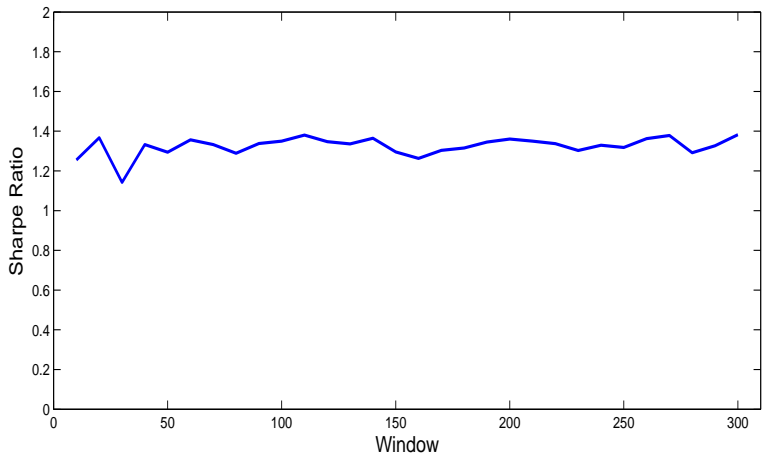}}
\caption{Sharp ratio sensitivity w.r.t w}
\label{fig:sensitivityWSharpe}
\end{center}
\vskip -0.2in
\end{figure}

While these results clearly provide a compelling proof of concept 
for the effectiveness of $\EP$, we chose $\lambda= 0.1$ in hindsight, which
affects both the return and the Sharpe ratio of the algorithm.
Is it possible to calibrate $\lambda$ online?
 To this end we employed a standard walk-forward procedure whereby 
$\lambda$ was sequentially optimized over a sliding window of size $w$ periods 
 so as to improve the Sharpe ratio. Although this routine eliminates the need to choose $\lambda$, 
 it introduces $w$ as a new hyper-parameter. Figure~\ref{fig:sensitivityWSharpe}
 depicts the sensitivity of the overall Sharpe ratio with respect to the window size $w \in [10, 300]$.
 It is evident that this procedure is not sensitive to choices of $w$ in this range.
$\EP$'s improvement of the Sharpe ratio  for the two base-algorithms settings is shown in 
 Table~\ref{Table:HPS}  (under $\lambda_{WF}$). The cumulative return of $\EP$ with sequentially calibrated 
 $\lambda$ appears in Figure~\ref{fig:EPvsORSAD}.

 \section{Conclusions}

In this paper we show how to effectively use the
$\ell_{\infty} / \ell_1$ regularization to improve risk-adjusted return performance.
While the original work introducing this norm \cite{JohnsonB2015} considered portfolios  on the stocks themselves, 
we propose to incorporate 
 this structured norm within a Newton style optimization and apply the learning algorithm
 over trading strategies based on a known partition of the stocks into industry sectors.
 Along the way, we also online optimize the choice of hyper-parameters and/or the choice of
  underlying trading algorithms. Our preliminary empirical study indicates that the proposed procedure
 can achieve a substantial improvement in the Sharpe ratio relative to the base algorithms themselves, relative to their 
 ensemble using the technique  of \Gcite{DasB2011}. Moreover, it substantially improves the original 
 method of  \Gcite{JohnsonB2015}. Further empirical evidence with qualitatively similar conclusions
 has been established and will be reported in the long version of this paper.
 
\Gcite{JohnsonB2015} attempted to dynamically compute sector 
groupings based on correlations. This  idea is very attractive
in the absence of prior knowledge or in fast changing markets,
and can be further extended to explore hierarchical structures in the stock market as discussed, e.g., in \Gcite{Mantegna1999}.
However, the regret guarantee in   \Gcite{JohnsonB2015}  cannot hold in its current form
using a dynamically changing grouping of the stocks. It would be very interesting
to extend their learning algorithm or ours to capture dynamically changing groupings. While the proposed procedure allows for improved Sharpe-ratios, it is still a ``long-only'' strategy, which  does not utilize short selling. While attractive in certain regulated settings like mutual funds, it cannot achieve market neutrality.
It would be very interesting to extend our technique so as to be market neutral.
As a final caveat, we must emphasize that
we have focused on an idealized ``frictionless'' setting that excludes
various elements such as
commissions, slippage and market impact.
While this setting is a reasonable starting point for considering risk reduction,these elements must be considered in  real-life applications.


 \appendix

 \bibliographystyle{named}
 \bibliography{ijcai16}

\begin{thebibliography}{}

\bibitem[\protect\citeauthoryear{Agarwal \bgroup \em et al.\egroup
  }{2006}]{AgarwalHKS2006}
A.~Agarwal, E.~Hazan, S.~Kale, and R.E. Schapire.
\newblock Algorithms for portfolio management based on the newton method.
\newblock In {\em Proceedings of the 23rd International Conference on Machine
  Learning}, pages 9--16. ACM, 2006.

\bibitem[\protect\citeauthoryear{Beck and Teboulle}{2003}]{BeckT2003}
A.~Beck and M.~Teboulle.
\newblock Mirror descent and nonlinear projected subgradient methods for convex
  optimization.
\newblock {\em Operations Research Letters}, 31(3):167--175, 2003.

\bibitem[\protect\citeauthoryear{Blum and Kalai}{1999}]{BlumK1999}
A.~Blum and A.~Kalai.
\newblock Universal portfolios with and without transaction costs.
\newblock {\em Machine Learning}, 35(3):193--205, 1999.

\bibitem[\protect\citeauthoryear{Borodin and El-Yaniv}{1998}]{Borodin2005}
A.~Borodin and R.~El-Yaniv.
\newblock {\em Online Computation and Competitive Analysis}.
\newblock Cambridge University Press, 1998.

\bibitem[\protect\citeauthoryear{Borodin \bgroup \em et al.\egroup
  }{2000}]{BorodinEG2000}
A.~Borodin, R.~El-Yaniv, and V.~Gogan.
\newblock On the tompetitive theory and practice of portfolio selection.
\newblock In {\em LATIN 2000: Theoretical Informatics}, pages 173--196. 2000.

\bibitem[\protect\citeauthoryear{Borodin \bgroup \em et al.\egroup
  }{2004}]{BorodinETG2004}
A.~Borodin, R.~El-Yaniv, and V.~Gogan.
\newblock Can we learn to beat the best stock?
\newblock {\em Journal of Artificial Intelligence Research}, pages 579--594,
  2004.

\bibitem[\protect\citeauthoryear{Cavaglia \bgroup \em et al.\egroup
  }{2000}]{CavagliaBA2000}
S.~Cavaglia, C.~Brightman, and M.~Aked.
\newblock The increasing importance of industry factors.
\newblock {\em Financial Analysts Journal}, 56(5):41--54, 2000.

\bibitem[\protect\citeauthoryear{Cover and Ordentlich}{1996}]{CoverO1996}
T.M. Cover and E.~Ordentlich.
\newblock Universal portfolios with side information.
\newblock {\em IEEE Transactions on Information Theory}, 42(2):348--363, 1996.

\bibitem[\protect\citeauthoryear{Cover}{1991}]{Cover1991}
T.M. Cover.
\newblock Universal portfolios.
\newblock {\em Mathematical Finance}, 1(1):1--29, 1991.

\bibitem[\protect\citeauthoryear{Das and Banerjee}{2011}]{DasB2011}
P.~Das and A.~Banerjee.
\newblock Meta optimization and its application to portfolio selection.
\newblock In {\em Proceedings of the 17th ACM SIGKDD International Conference
  on Knowledge Discovery and Data Mining}, pages 1163--1171. ACM, 2011.

\bibitem[\protect\citeauthoryear{Duchi \bgroup \em et al.\egroup
  }{2010}]{DuchiSST2010}
J.C. Duchi, S.~Shalev-Shwartz, Y.~Singer, and A~.Tewari.
\newblock Composite objective mirror descent.
\newblock In {\em COLT}, pages 14--26, 2010.

\bibitem[\protect\citeauthoryear{Even-Dar \bgroup \em et al.\egroup
  }{2006}]{EvenKW2006}
E.~Even-Dar, M.~Kearns, and J.~Wortman.
\newblock Risk-sensitive online learning.
\newblock In {\em Algorithmic Learning Theory}, pages 199--213. Springer, 2006.

\bibitem[\protect\citeauthoryear{Grinold \bgroup \em et al.\egroup
  }{1989}]{Grinold1989}
R.C. Grinold, A.~Rudd, and D.~Stefek.
\newblock Global factors: fact or fiction?
\newblock {\em The Journal of Portfolio Management}, 16(1):79--88, 1989.

\bibitem[\protect\citeauthoryear{Harlow}{1991}]{Harlow1991}
W.V. Harlow.
\newblock Asset allocation in a downside-risk framework.
\newblock {\em Financial Analysts Journal}, 47(5):28--40, 1991.

\bibitem[\protect\citeauthoryear{Hazan \bgroup \em et al.\egroup
  }{2007}]{AK2007}
E.~Hazan, A.~Agarwal, and S.~Kale.
\newblock Logarithmic regret algorithms for online convex optimization.
\newblock {\em Machine Learning}, 69(2-3):169--192, 2007.

\bibitem[\protect\citeauthoryear{Helmbold \bgroup \em et al.\egroup
  }{1998}]{helmboldSSW1998}
D.P. Helmbold, R.E. Schapire, Y.~Singer, and M.K. Warmuth.
\newblock On-line portfolio selection using multiplicative updates.
\newblock {\em Mathematical Finance}, 8(4):325--347, 1998.

\bibitem[\protect\citeauthoryear{Johnson and Banerjee}{2015}]{JohnsonB2015}
N.~Johnson and A.~Banerjee.
\newblock Online resource allocation with structured diversification.
\newblock 2015.

\bibitem[\protect\citeauthoryear{Li and Hoi}{2012}]{LiH2012}
B.~Li and S.C.H. Hoi.
\newblock On-line portfolio selection with moving average reversion.
\newblock In {\em Proceedings of the 29th International Conference on Machine
  Learning (ICML-12)}, pages 273--280, 2012.

\bibitem[\protect\citeauthoryear{Li and Hoi}{2014}]{LiH2014}
B.~Li and S.C.H. Hoi.
\newblock Online portfolio selection: A survey.
\newblock {\em ACM Computing Surveys (CSUR)}, 46(3):35, 2014.

\bibitem[\protect\citeauthoryear{Li \bgroup \em et al.\egroup }{2015}]{OLPS}
B.~Li, D.~Sahoo, and S.C.H. Hoi.
\newblock Olps: A toolbox for online portfolio selection.
\newblock {\em Journal of Machine Learning Research (JMLR)}, 2015.

\bibitem[\protect\citeauthoryear{Mantegna}{1999}]{Mantegna1999}
R.N. Mantegna.
\newblock Hierarchical structure in financial markets.
\newblock {\em The European Physical Journal B-Condensed Matter and Complex
  Systems}, 11(1):193--197, 1999.

\bibitem[\protect\citeauthoryear{Markowitz}{1952}]{Markowitz1952}
H.~Markowitz.
\newblock Portfolio selection*.
\newblock {\em The journal of finance}, 7(1):77--91, 1952.

\bibitem[\protect\citeauthoryear{Nemirovsky and Yudin}{1985}]{NemirovskyY1983}
A.S. Nemirovsky and D.B. Yudin.
\newblock Problem complexity and method efficiency in optimization.
\newblock {\em SIAM Review}, 27(2):264--265, 1985.

\bibitem[\protect\citeauthoryear{Ordentlich and Cover}{1996}]{OrdentlichC1996}
E.~Ordentlich and T.M. Cover.
\newblock On-line portfolio selection.
\newblock In {\em Proceedings of the 9th Annual Conference on Computational
  Learning Theory}, COLT, pages 310--313, 1996.

\bibitem[\protect\citeauthoryear{Pardo}{1992}]{Pardo1992}
R.~Pardo.
\newblock {\em Design, testing, and optimization of trading systems}.
\newblock John Wiley \& Sons, 1992.

\bibitem[\protect\citeauthoryear{Sharpe}{1964}]{Sharpe1964}
W.F. Sharpe.
\newblock Capital asset prices: A theory of market equilibrium under conditions
  of risk.
\newblock {\em The Journal of Finance}, 19(3):425--442, 1964.

\bibitem[\protect\citeauthoryear{Sharpe}{1966}]{Sharpe1966}
W.F. Sharpe.
\newblock Mutual fund performance.
\newblock {\em Journal of business}, pages 119--138, 1966.

\bibitem[\protect\citeauthoryear{Shen \bgroup \em et al.\egroup
  }{2015}]{ShenWJZ2015}
W.~Shen, J.~Wang, Y.G. Jiang, and H.~Zha.
\newblock Portfolio choices with orthogonal bandit learning.
\newblock In {\em Proceedings of the 24th International Conference on
  Artificial Intelligence}, pages 974--980. AAAI Press, 2015.

\bibitem[\protect\citeauthoryear{Solnik}{1995}]{Solnik1995}
B.H. Solnik.
\newblock Why not diversify internationally rather than domestically?
\newblock {\em Financial Analysts Journal}, 51(1):89--94, 1995.

\bibitem[\protect\citeauthoryear{Yuan and Lin}{2006}]{Yuan2006}
M.~Yuan and Y.~Lin.
\newblock Model selection and estimation in regression with grouped variables.
\newblock {\em Journal of the Royal Statistical Society: Series B (Statistical
  Methodology)}, 68(1):49--67, 2006.

\bibitem[\protect\citeauthoryear{Zinkevich}{2003}]{Zinkevich2003}
M.~Zinkevich.
\newblock Online convex programming and generalized infinitesimal gradient
  ascent.
\newblock In {\em Machine Learning, Proceedings of the 20th International
  Conference ({ICML})}, pages 928--936, 2003.

\end{thebibliography}
 
\end{document}